\documentclass[9pt,a4paper,english]{article}
\usepackage[top=1.5in,left=1.5in,right=1.5in,footskip=0.4in,marginparwidth=2in]{geometry}

\usepackage{graphicx}
\usepackage{multirow}
\usepackage{amsmath}
\usepackage{mathrsfs}%
\usepackage{xcolor}%
\usepackage{textcomp}%
\usepackage{tablefootnote}
\usepackage{booktabs}
\usepackage{algorithm}
\usepackage{algpseudocode}
\usepackage{listings}
\usepackage{url}
\usepackage{float}
\usepackage{tikz}
\usepackage{pifont}
\usetikzlibrary{shapes.geometric, arrows}
\usepackage{booktabs,subcaption,dcolumn}
\usepackage{subcaption}
\usepackage[graphicx]{realboxes}
\usepackage{amsbsy}
\usepackage{wasysym }
\usepackage{enumitem}
\usepackage{titlesec}
\newcommand{\wrap}[1]{\begin{tabular}{@{}c@{}}#1\end{tabular}}

\begin{document}

\newcommand{\footremember}[2]{%
    \footnote{#2}
    \newcounter{#1}
    \setcounter{#1}{\value{footnote}}%
}
\newcommand{\footrecall}[1]{%
    \footnotemark[\value{#1}]%
} 
\title{Exploring the Decentraland Economy: Multifaceted Parcel Attributes, Key Insights, and Benchmarking}
\author{%
  Dipika Jha\footremember{*}{Corresponding Author: dipika\_23ps22@iitp.ac.in}%
  \and Ankit K. Bhagat
  \and Raju Halder
  \and Rajendra N. Paramanik
  \and Chandra M. Kumar
  }

\date{Indian Institute of Technology Patna, India}

\maketitle

\begin{abstract}
This paper presents a comprehensive Decentraland parcels dataset, called \textsf{IITP-VDLand}, sourced from diverse platforms such as \textit{Decentraland}, \textit{OpenSea}, \textit{Etherscan}, \textit{Google BigQuery}, and various \textit{Social Media Platforms}. Unlike existing datasets which have limited attributes and records, \textsf{IITP-VDLand} offers a rich array of attributes, encompassing parcel characteristics, trading history, past activities, transactions, and social media interactions. Alongside, we introduce a key attribute in the dataset, namely Rarity score, which measures the uniqueness of each parcel within the virtual world. Addressing the significant challenge posed by the dispersed nature of this data across various sources, we employ a systematic approach, utilizing both available APIs and custom scripts, to gather it. Subsequently, we meticulously curate and organize the information into four distinct fragments: (1) Characteristics, (2) OpenSea Trading History, (3) Ethereum Activity Transactions, and (4) Social Media. We envisage that this dataset would serve as a robust resource for training machine- and deep-learning models specifically designed to address real-world challenges within the domain of Decentraland parcels. The performance benchmarking of more than 20 state-of-the-art price prediction models on our dataset yields promising results, achieving a maximum $R^2$ score of 0.8251 and an accuracy of 74.23\% in case of Extra Trees Regressor and Classifier. The key findings reveal that the ensemble models perform better than both deep learning and linear models for our dataset. We observe a significant impact of coordinates, geographical proximity, rarity score, and few other economic indicators on the prediction of parcel prices.
\end{abstract}

\section{Introduction}
Embarking on an innovative journey in 2017 by the visionary founders, Esteban Ordano and Ari Meilich, Decentraland \cite{Decentraland} has emerged as one of the most popular Metaverse platform in the present day. This pioneering venture allows users to be part of a shared digital experience where they can play games, socialize, create, buy, or sell digital items. Few of the key components of Decentraland include LAND (virtual real estate), Parcels (individual units of LAND), Estates (collection of parcels), Avatars (virtual user representations), and the builder-tool for creating 3D content. Notably, the popularity of Decentraland surged when virtual parcels were traded for almost \$1 million in June 2021, followed by a notable transaction of cryptocurrency worth \$2.4 million, with the buyer being the crypto investor Tokens.com. As reported in \cite{DAUReport}, in 2022, Decentraland experienced around 8000 daily active users with large number of repeat visitors to explore and spend time on the platform. This series of events played a crucial role in propelling the market capitalization to a noteworthy \$903,847,923 in the recent year \cite{Marketcap}. Unlike other Metaverse platforms, Decentraland is governed by the DAO, a Decentralized Autonomous Organization, that utilizes off-chain voting for community engagement.  

As the Decentraland experiences a surge in popularity and an expanding user-base, a plethora of interesting research avenues emerge. Among these, notable research problems include predicting parcel prices and forecasting their future values, classifying high-valued parcels, and tracking intricate trends within the Decentraland ecosystem. Recent research efforts in the literature aimed at understanding the impact of various features on the dynamic pricing behavior of Decentraland Non-Fungible Tokens (NFTs) \cite{dowling2022fertile, goldberg2021land, yencha2023spatial, nadini2021mapping, dowling2022non, schonbaum2022decentraland, pinto2022nft, kaneko2021time, luo2023understanding, costa2023show,brunet2023exploring, luo2023unveiling}. 

However, the datasets used in their training phases contain a limited number of records and attributes of Decentraland parcels, therefore lacking the comprehensive representation necessary for analyzing the multifaceted influences on parcel dynamics within Decentraland. Specifically, out of the four publicly available datasets \cite{nadini2021mapping, luo2023understanding, Dataset3, Dataset4}, two contain a maximum of 6,141 records with 37 attributes, while the other two  contains social media comments/discussions. This underscores the pressing need for a comprehensive dataset encompassing Decentraland parcels along with their detailed attributes, which are currently fragmented and scattered across diverse sources.

To this aim, this paper is dedicated to the development of a comprehensive dataset, called \textsf{IITP-VDLand}, tailored specifically for the virtual realm of Decentraland parcels. Our extensive compilation encompasses a diverse array of attributes, spanning parcels' characteristics and their trading history, as well as transaction-specific details recorded on the Ethereum blockchain including social media data. Drawing from various sources such as \textit{Decentraland} \cite{Decentraland}, \textit{OpenSea} \cite{OpenSea}, \textit{Etherscan} \cite{Etherscan}, \textit{Google BigQuery} \cite{BigQuery}, \textit{Discord} \cite{Discord}, \textit{Telegram} \cite{Telegram}, and              \textit{Reddit} \cite{Reddit}, this dataset serves as a robust resource for training machine- and deep-learning models specifically designed to address real-world challenges within the domain of Decentraland parcels.

In particular, our dataset comprises in total 92,598 parcels with 81 attributes. These parcels are associated with various characteristics attributes including their coordinates, geographical proximity from prominent locations, visual link, unique Id, and rarity score. This is worth emphasizing that, for the first time, we use one of the most popular rarity meters, namely Rarity.tools, to calculate the rarity score of the parcels, which measures the uniqueness of each parcel in comparison to others. To capture the market dynamics of Decentraland, we leverage the extensive trading history available on OpenSea, the largest secondary marketplace. This allows us to access offers prices and sales prices for each parcel over a specified time period. Since Decentraland operates on the Ethereum blockchain, to delve into the specifics of Ethereum on-chain transactions, we aim to include transaction details such as gas prices, bidding history, and parcel-related activities. These attributes encompass details like invoked methods, block-specific information, mining specifics, and more. In essence, we seek to comprehensively analyze Ethereum transactions within Decentraland by capturing a broad spectrum of relevant data points, facilitating deeper insights into user interactions and the ecosystem's dynamics. We also dive into social media platforms to get a gist of people's views and sentiments. These distinctive attributes significantly enrich our dataset, offering a novel perspective for research and academic exploration. By delving into the economic dimension of Decentraland, researchers can gain a profound understanding of this virtual world. To enhance clarity, we've organized our dataset into four fragments: (1) Characteristics Data-Fragment, (2) OpenSea Trading History Data-Fragment, (3) Ethereum Activity Transactions Data-Fragment, and (4) Social Media Data-Fragment. 

To the best of our knowledge, this is the first benchmark dataset that is in-depth and tailor-made for various machine- and deep-learning applications of Decentraland parcels. Unlike existing Decentraland NFT datasets which are notably incomplete and lack critical attributes, \textsf{IITP-VDLand} sets a new standard encompassing details ranging from parcels' metadata, transactional details to social media data. To summarize, the major contributions of our paper are:
\begin{itemize}
    \item We introduce \textsf{IITP-VDLand}, an extensive dataset comprising Decentraland's parcel NFTs sourced from diverse platforms, including \textit{Decentraland} \cite{Decentraland}, \textit{OpenSea} \cite{OpenSea}, \textit{Etherscan} \cite{Etherscan}, \textit{Google BigQuery} \cite{BigQuery}, \textit{Discord} \cite{Discord}, \textit{Telegram} \cite{Telegram}, and \textit{Reddit} \cite{Reddit}. This dataset serves as a robust resource for training machine- and deep-learning models tailored to tackle real-world challenges within the Decentraland NFT domain. The dataset is made available for download at: \url{https://dx.doi.org/10.21227/qv8s-7n53}
    
    \item We meticulously curate the data, creating four distinct fragments: (1) Characteristics Data-Fragment, (2) OpenSea Trading History Data-Fragment, (3) Ethereum Activity Transactions Data-Fragment, and (4) Social Media Data-Fragment. Alongside the existing data, we also introduce a significant attribute of Decentraland parcels in the dataset, namely Rarity score, which quantifies the uniqueness of each parcel within the virtual world.

    \item We conduct a performance benchmarking of more than 20 state-of-the-art price prediction models on our novel dataset, aiming at assessing their predictive performance and determine which models are best suited for our dataset. The results demonstrate that ensemble models perform better than both deep learning and linear models for our dataset, achieving highest $R^2$ score of 0.8251 and accuracy of 74.23\% in case of Extra Trees Regressor and Classifier. Furthermore, the ablation study reveals the importance of various attributes present in our dataset, demonstrating the substantial influence of coordinates, geographical proximity, rarity score, and few other economic indicators on parcel price prediction.
    
\end{itemize}

\noindent The structure of the rest of the paper is organized as follows: In Section \ref{sec:Decentraland}, we provide a preliminary background about Decentraland Metaverse platform. In Section \ref{sec: overview}, we present an overview of previous research endeavors and publicly available datasets in the literature. A detailed description of our proposed dataset is provided in Section \ref{sec:ourDataset}, emphasizing its composition, sourcing methodology, and inherent attributes. Section \ref{eda_app} conducts an exploratory data analysis, highlighting key insights and patterns in our dataset. Section \ref{sec:benchmark} outlines the empirical results from the performance benchmarking of our dataset using over 20 state-of-the-art price prediction models. Finally, in Section \ref{sec:Discussion}, we discuss our key findings and highlight the potential future avenues.

\section{Decentraland: A Quick Tour}
\label{sec:Decentraland}

Decentraland \cite{Decentraland} is a Metaverse built on the Ethereum blockchain. Users can buy, sell, and develop virtual real estate called LAND, which is divided into unique NFT parcels. This allows users creative freedom to build virtual businesses, events, and experiences. Beyond parcel ownership, Decentraland aims to simplify development with user-friendly tools while also empowering users to earn passive income by renting out their virtual property.

\begin{figure*}[t]
    \centering
    \subfloat[Decentraland map as bird eye view]{\label{fig:birdview}
        \includegraphics[width=0.5\textwidth]{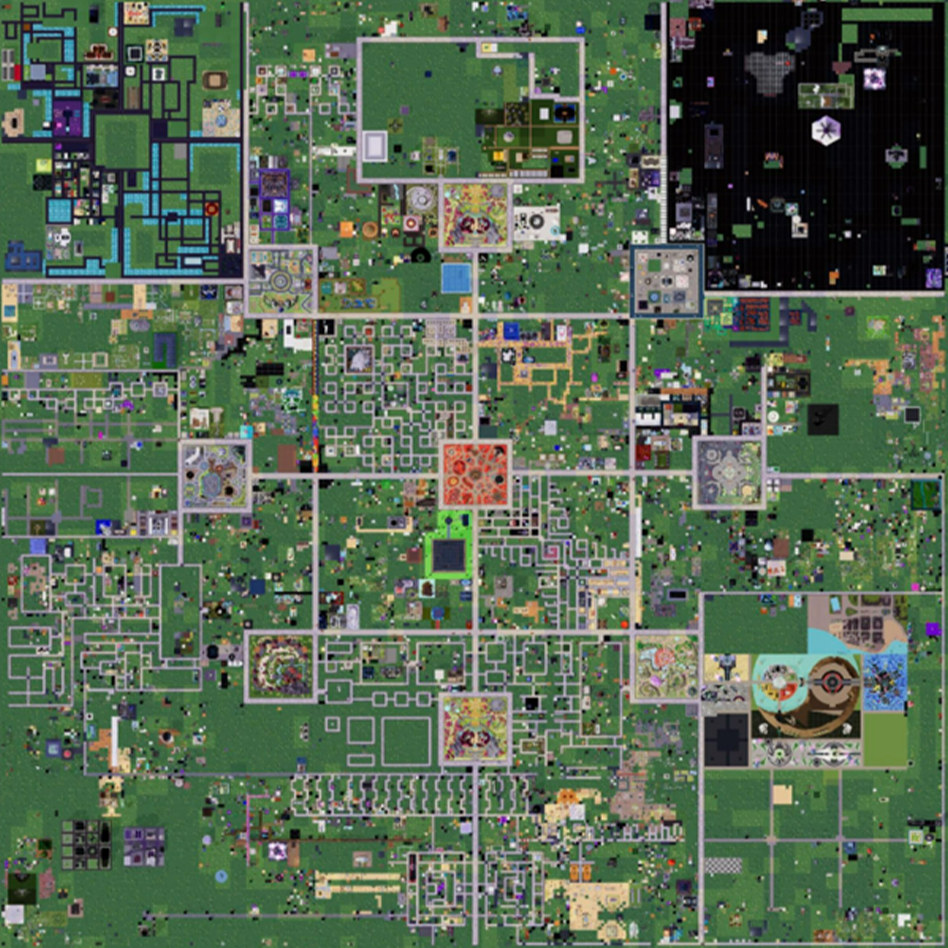}
    }
    % \hfill
    \subfloat[Decentraland map of diverse landscape]{\label{fig:decentralandElements}
        \includegraphics[width=0.5\textwidth]{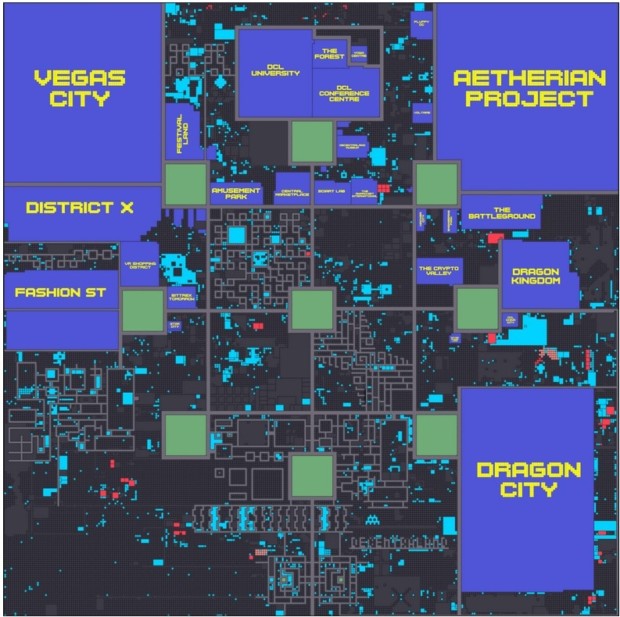}
    }
    \caption{Decentraland Map}
    \label{fig:decentralandmap}
\end{figure*}

Decentraland is governed by 126 smart contracts, with 101 on Ethereum and 25 on Polygon. It utilizes two tokens: LAND and MANA. LAND is a non-fungible token representing finite 3D virtual spaces within Decentraland. MANA, a fungible token based on the ERC20 standard, is the platform's cryptocurrency used for purchasing LAND parcels and accessing various services. LAND tokens adhere to the ERC721 standard. Decentraland is divided into 92,598 parcels as shown in the Decentraland Map in Figure \ref{fig:decentralandmap}, each measuring 16m x 16m (52 × 52 ft) and identified by unique cartesian coordinates (x, y) within the range of (-150 to 163) and (-150 to 158), respectively. The Decentraland Map \cite{NFTPlaza} shows location of all the parcels including both private owners and themed community districts. Privately owned Parcels (Dark Grey), are available for trading on the marketplace. DISTRICTS (Purple), are privately held and not available for sale. PLAZAS (Large Green Squares) are non-sellable areas with high foot traffic where players respawn. Finally, ROADS (Light Grey, Straight Lines) non-sellable spaces.

An association of two or more directly adjacent parcels is called Estate. These parcels must be directly adjacent and can not be separated by a road, plaza or another parcel. Currently, 2,645 estates exist within Decentraland's diverse landscape of districts, plazas, roads, and parcels, as depicted in Figure \ref{fig:decentralandmap}. Unlike conventional web domains, Decentraland parcels are unique for their distinctive adjacency feature. This contiguity enables spatial content exploration and the formation of themed districts. The finite nature of parcel adjacencies, compared to the limitless hyperlinks of the web, provides content visibility from neighboring parcels within a certain distance. This structure encourages the creation of districts attracting targeted traffic, offering immersive themed experiences. The scarcity of parcels compels strategic acquisition in high-traffic areas, fostering secondary markets for ownership and rentals.

Decentraland parcels gain value through their proximity to popular areas, application hosting capabilities, and function as identity mechanisms. These factors drive developers and creators to acquire parcels for building experiences and engaging audiences. Although unclaimed parcels share a standard exchange rate (1000 MANA = 1 Parcel), each possesses unique characteristics that influence their price on the secondary market. This price variability stems from differences in factors like adjacency to desirable areas and traffic, highlighting the individual appeal of each parcel \cite{White_Paper}.
Parcel and estate transactions in Decentraland occur on the Ethereum mainnet, facilitated by various smart contracts like LANDRegistry, MANAToken, and others. Notably, the Marketplace smart contract on Ethereum manages the sale of parcels, estates, names, and wearables, while a separate Polygon-based contract handles the trading of wearables and emotes.

\section{Overview of Prior Efforts}\label{sec: overview}
There have been number of attempts to solve various research problem related to Decentraland. In this section, we first provide an overview of the existing research efforts addressing various applicative aspects of Decentraland. Then we discuss the details about the Decentraland datasets available so far and their utilities.

\subsection{Literature Review}\label{LiteratureReview}
Mitchael Dowling in his early study \cite{dowling2022fertile} highlighted the pricing behaviour of Decentraland Land NFTs. Interestingly, it is observed that, despite of having a discrepancy of the pricing behaviour in the early stage of market, the value of NFT was continuously on rise. Later, Goldberg et al. \cite{goldberg2021land} identified that the pricing of the NFTs, which has not been considered in \cite{dowling2022fertile}, is closely linked with the location of the parcels, particularly the locations close to the center part or have some memorable addresses tend to have higher price. Further, Christopher Yencha \cite{yencha2023spatial} considered spatial characteristics of Decentraland parcels to provide an evidence of the effects of locations on the price, similar to the real estate market. To map the market trends and the trade network after NFT market experienced a record sale in 2021, Nadini et al. \cite{nadini2021mapping} used statistical properties of the market and showed how visual features and sales history are good predictors for the price of Decentraland as well as other NFTs. As cryptocurrency pricing behaviour is an important factor for the market of NFTs, the authors in \cite{dowling2022non} presented a wavelet coherence analysis between NFT pricing and cryptocurency. Observably, a spillover index showed only limited volatility transmission effect between cryptocurrency and NFTs. Rachel Schonbaum in \cite{schonbaum2022decentraland} envisioned the possibility of improved market efficiency in Decentraland’s LAND market, which in turn has positive implications for NFT usage on the Metaverse. According to them, the NFT-world is likely to change in the next decade, bringing a lot of potential for positive changes in the near future.
 
Few significant research attempts demonstrating the impact of social media on the pricing of popular NFT's, including Decentraland is reported in \cite{pinto2022nft,kaneko2021time,luo2023understanding}. Christian Pinto-Gutiérrez et al. \cite{pinto2022nft} showed how the change in cryptocurrency market (Bitcoin and Ether) draw attention of the investors of NFT's, such as Decentraland and Cryptopunk, by examining the data of Google search trends. A time series analysis on Google trends data of cryptocurrency prices for NFT's is presented in \cite{kaneko2021time}. Junliang Luo et al. \cite{luo2023understanding} performed tweet keyword analysis based on Decentraland along with other top 18 NFT projects, demonstrating the impact of feature word extraction to influence the price of NFTs.

To predict NFT performance by solely relying on images and description, the authors in \cite{costa2023show} proposed a multimodal representation based learning by considering Decentraland transactions data along with other NFTs. S.C.Brunet et al. \cite{brunet2023exploring} provided a detailed study to show how the factors related to the revenue-generating potential of the Decentraland parcels are more likely to play a role in its pricing, rather the user traffic on the parcel. J.Luo et al. \cite{luo2023unveiling} analysed user behavior in Decentraland using graph analysis technique, along with an examination of the user traffic and transactions. This is observed that interaction in virtual world is different from the traditional social-media platform. Majority of the interaction is based on economic incentive, rather than socializing. Some people even may not wish to join due to the lack of knowledge about blockchain technology. 

\subsection{Available Dataset}
\label{AvailableDataset}

Let us mention below the list of publicly available Decentraland datasets.
\begin{itemize}
     \item \textit{Dataset-1} \cite{nadini2021mapping}: Nadini et al. created a dataset to give a general overview of NFT market, consisting of 6.1 million trades of 4.7 million NFTs in 160 cryptocurrencies, covering the period between June 23, 2017 and April 27, 2021. The dataset was primarily obtained from Ethereum and WAX blockchains, using various open source APIs: Cryptokitties sales, Gods-unchained, Decentraland, OpenSea and Atomic API. They considered different NFTs and grouped them in six categories, i.e Art, Collectible, Games, Metaverse, Utility and Other. Decentraland was one of the NFT to be grouped in Metaverse category. The Decentraland dataset contains total 16,944 records categorized as: ens- 839, estate- 1,368, parcel- 6,141 and wearable- 8,645, sold in between June 2017- April 2021.
    
     \item \textit{Dataset-2} \cite{luo2023understanding}: Unlike previous dataset, Luo et al. explored the relationship between the NFT social media communities and the NFT price in term of the tweet number and the content of the tweets. They collected top 19 NFT token trade transactions and tweets, one of them was the Decentraland transactions data with 2,09,737 records, publicly available in Google BigQuery.

    \item \textit{Dataset-3} \cite{Dataset3}: This is Decentraland NFT Virtual Estate dataset including NFT land price and features, which is publicly available on Kaggle. The data is categorized into various estates according to their features , limited to only 1,999 records with highest sales count and less number feature for a perfect analysis. Kaggle has not mentioned about any study that uses this dataset for analysis.  

    \item \textit{Dataset-4} \cite{Dataset4}: This is a social media dataset which contains total 3,481 users' comments about Decentraland NFTs on Reddit and is publicly available on Kaggle. Even though this dataset is not being used by any researcher yet, we believe that it would be appropriate for analyzing the impact of social media on the price of Decentraland NFTs. 
\end{itemize}

\noindent Table \ref{tab:my-table} depicts a comparative summary of these existing related datasets and our proposed \textsf{IITP-VDLand} dataset. 

\begin{table*}[t]
\centering\scriptsize
\subfloat[Comparisons w.r.t. data sources and number of records] {\label{}
\begin{tabular}{|p{2cm}|p{2.1cm}|p{4cm}|c|}
\hline
\textbf{Datasets} & \textbf{Source(s)} & \textbf{Record(s)} & \textbf{Attributes Count}\\\hline
\textbf{Dataset-1 \cite{nadini2021mapping}} & Decentraland & Decentraland Parcel: 6141 & 37 \\\hline
\textbf{Dataset-2 \cite{luo2023understanding}} & \begin{tabular}[c]{@{}p{2cm}@{}}Google BigQuery,\\ Twitter\end{tabular} & \begin{tabular}[c]{@{}p{2cm}@{}}Decentraland Tweets: 51692\\ Decentraland Transactions: 209737\end{tabular} & 11 \\\hline 
\textbf{Dataset-3 \cite{Dataset3}} & \begin{tabular}[c]{@{}p{2cm}@{}}Ethereum Blockchain,\\ Decentraland\end{tabular} & Decentraland Parcel: 1999 & 14 \\\hline
\textbf{Dataset-4 \cite{Dataset4}} & \begin{tabular}[c]{@{}p{2cm}@{}}Reddit,\\ Pushshift\end{tabular}  & Comments: 3481 & - \\\hline

\textbf{IITP-VDLand (Proposed)}  & \begin{tabular}[c]{@{}p{4cm}@{}}Decentraland,\\ OpenSea,\\ Google BigQuery,\\ Etherscan,\\ Discord,\\ Telegram,\\ Reddit\end{tabular} & \begin{tabular}[c]{@{}p{4cm}@{}}Parcel's Characteristics: 92598\\ Parcel's OpenSea Sales: 20092\\ Parcel's OpenSea Offers: 2246545\\ Parcel's Ethereum Bidding: 40195\\ Parcel's Ethereum Sales: 20092\\ Parcel's Other Activity Transactions: 202136\\ Social Media: 317801\end{tabular} & 81 \\\hline
\end{tabular}}
\vspace{0.2in}
\subfloat[Comparisons w.r.t. data attributes (Note: {\LEFTcircle}= Partially Available, {\ding{51}}= Available, {\ding{53}} = Not Available, - = Not mentioned)]{\label{}
\begin{tabular}{|p{3cm}|c|c|c|c|c|}
\hline
\bf{Attribute(s)} & \wrap{\textbf{Dataset-1}\\ \cite{nadini2021mapping}} & \wrap{\textbf{Dataset-2}\\ \cite{luo2023understanding}} & \wrap{\textbf{Dataset-3}\\ \cite{Dataset3}} & \wrap{\textbf{Dataset-4}\\ \cite{Dataset4}} & \wrap{\textbf{IITP-VDLand}\\ (Proposed)} \\ \hline
Geographical Proximity & \LEFTcircle & \ding{55} & \LEFTcircle & \ding{55} & \ding{51} \\ \hline
Rarity Score & \ding{55} & \ding{55} & \ding{55} & \ding{55} & \ding{51} \\ \hline
Visual Link & \ding{51} & \ding{55} & \ding{55} & \ding{55} & \ding{51} \\ \hline
Sales Data & \ding{51} & \ding{55} & \LEFTcircle & \ding{55} & \ding{51} \\ \hline
Offers Data & \ding{55} & \ding{55} & \LEFTcircle & \ding{55} &\ding{51}  \\ \hline
Ethereum Transactions Data & \ding{55} & \LEFTcircle & \ding{55} & \ding{55} & \ding{51} \\ \hline
Social Media Data & \ding{55} & \ding{51} & \ding{55} & \ding{51} & \ding{51} \\ \hline
\end{tabular}}
\caption{Comparison Between Existing and Proposed Datasets}
\label{tab:my-table}
\end{table*}

\subsection{Why New Dataset?}
Our dataset makes a valuable contribution to understanding blockchain technology and the NFT market, particularly Web3 and virtual land within Decentraland. As Decentraland's prominence has grown, it has attracted creators, investors, and researchers alike, all drawn to the unique opportunities offered by its blockchain-based virtual world.
Existing datasets, while valuable, often fall short in providing the detailed records and attributes necessary for a thorough analysis of the factors influencing parcel dynamics within Decentraland. Some focus solely on sales activities, others on social media influence, leaving a gap for a comprehensive resource. Table \ref{tab:my-table} highlights these limitations.
Our dataset addresses this gap by providing a comprehensive repository of information crucial for understanding the virtual land market, valuing assets, and analyzing investments. Spanning five years (2018-2023), it captures the full spectrum of market trends – peaks, downturns, and evolving patterns – offering insights into the NFT market's dynamics. 

Furthermore, our dataset adheres to the FAIR principles \cite{wilkinson2016fair}. It is Findable, publicly available through the resources mentioned earlier. It is Accessible, as it can be accessed globally and is provided in the widely used JSON format, ensuring broad compatibility. The use of JSON also makes our dataset Interoperable, as most programming languages support working with JSON data. Lastly, our dataset is Reusable, with detailed provenance to ensure transparency and traceability. However, assembling such a dataset poses significant challenges due to fragmented information scattered across diverse sources, making it difficult to obtain a holistic view of Decentraland's past and present.

Our study addresses this challenge by introducing the \textsf{IITP-VDLand} dataset, a novel and comprehensive resource dedicated to Decentraland's Parcel NFTs. As delineated in Table \ref{tab:my-table}, our dataset incorporates several pivotal features, including: (1) Rarity score: A metric quantifying the uniqueness of NFTs within a collection, influencing users' purchasing decisions. (2) Geographical proximity: Detailing the spatial relationships between parcels, providing insights into clustering and neighborhood dynamics. (3) Blockchain- and transaction-specific details: Offering a granular understanding of transaction histories and blockchain attributes, crucial for discerning patterns and behaviors. (4) Comprehensive activity tracking: Enabling a complete view of all transactions and activities within Decentraland parcels, facilitating deep-dive analyses into specific events and trends.

To enhance usability across diverse applications, we fragment the dataset according to various criteria, tailoring it to meet specific analytical needs. This fragmentation facilitates targeted analyses and promotes wider adoption across the Decentraland community and beyond. 

\begin{figure*}[t]
\centering
\includegraphics[width=\textwidth]{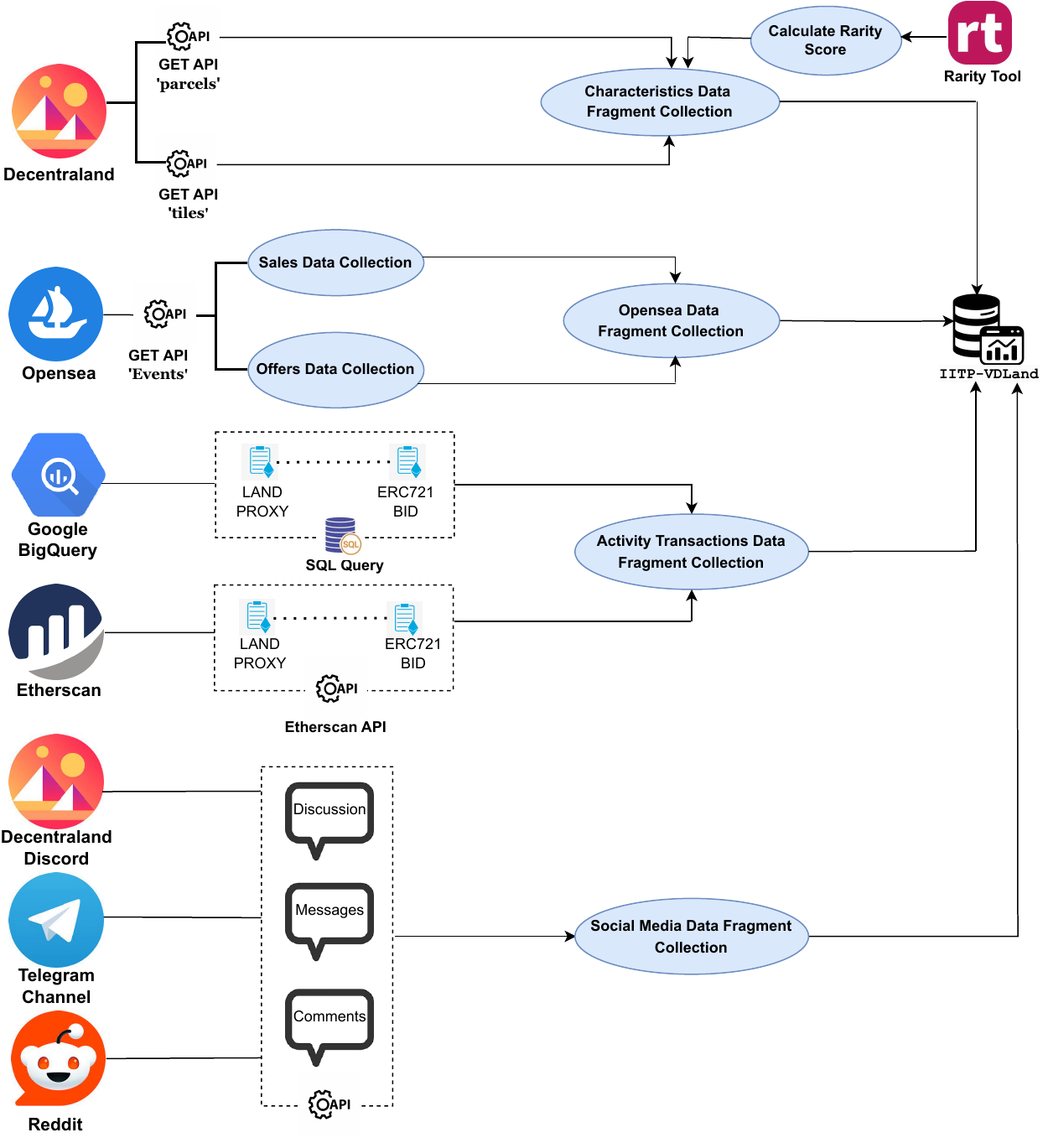}
\caption{Overview of Data collection process}\label{data_collection2}
\end{figure*}

\section{Our Dataset \textsf{IITP-VDLand}}\label{sec:ourDataset}
In this section, we present \textsf{IITP-VDLand}, an expansive dataset of Decentraland's \textit{parcel} NFTs, which are gathered from diverse sources such as \textit{Decentraland} \cite{Decentraland}, \textit{OpenSea} \cite{OpenSea}, \textit{Etherscan} \cite{Etherscan}, \textit{Google BigQuery} \cite{BigQuery}, \textit{Discord} \cite{Discord}, \textit{Telegram} \cite{Telegram}, and \textit{Reddit} \cite{Reddit}. This dataset serves as a robust resource for training machine- and deep-learning models specifically designed to address real-world challenges within the domain of Decentraland NFTs.

In our data collection process, we harness all possible relevant data sources and APIs to collect and compile our dataset. In particular, we obtain parcels metadata using \textit{Decentraland API} `\textit{tiles}' \cite{tiles} and `\textit{parcels}' \cite{parcels}\, which offers an up-to-date characteristics information, such as unique Id, name, coordinates, geographical proximity, rarity score, and visual link about each parcel. For a deeper understanding of lands' sales history and offers received, we seamlessly integrate the \textit{OpenSea API} `\textit{Get Events (by NFT)}' \cite{OpenSea}, which covers temporal perspectives. Tracking gas prices, a pivotal aspect in Ethereum transactions, is facilitated through \textit{Google BigQuery} \cite{BigQuery}, supplying information on Decentraland transactions' costs. Additionally, to extract essential details regarding the activities initiated through these transactions, we leverage \textit{Etherscan} \cite{Etherscan}, which contributes to a well-labeled dataset. Subsequently, we expand our data scope to prominent social media platforms, including \textit{Discord} \cite{Discord}, \textit{Telegram} \cite{Telegram}, and \textit{Reddit} \cite{Reddit}. We systematically collect individuals' comments and opinions pertaining to Decentraland providing invaluable insights within the online community. Our thorough data collection process ensures that \textsf{IITP-VDLand} is well-equipped to support diverse analytical and predictive endeavors within the Decentraland virtual ecosystem. A global overview of our data collection process is pictorially shown in Figure \ref{data_collection2}.

Our dataset comprises 92,598 parcels, serving as a rich source for analysis. Among them, 9,220 parcels are linked with historical sales data which creates a valuable time series encompassing 20,092 records. In connection with these parcels, we collect in total 2,62,423 records from both Google BigQuery and Etherscan, out of which 40,195 records represent bidding information, 20,092 records represent sales information, and 2,02,136 records represent other activities (such as minting, creating estate, claiming rewards, delisting parcels, etc.) over the Decentraland. Apart from these, we also collect 3,17,801 pieces of content from social media platforms that provide valuable information to predict the rise and fall of demands in parcel pricing. Furthermore, the dataset contains a substantial 22,46,545 records of offer information associated with these parcels, enabling an unrestricted examination of buying and selling activities within the OpenSea marketplace.  

According to the nature of information linked to the parcels, we divide our dataset into the following four fragments: (1) Characteristics Data-Fragment, (2) OpenSea Trading History Data-Fragment, (3) Ethereum Activity Transactions Data-Fragment, and (4) Social Media Data-Fragment.

\begin{figure*}[t]
\centering
\includegraphics[width=0.8\textwidth]{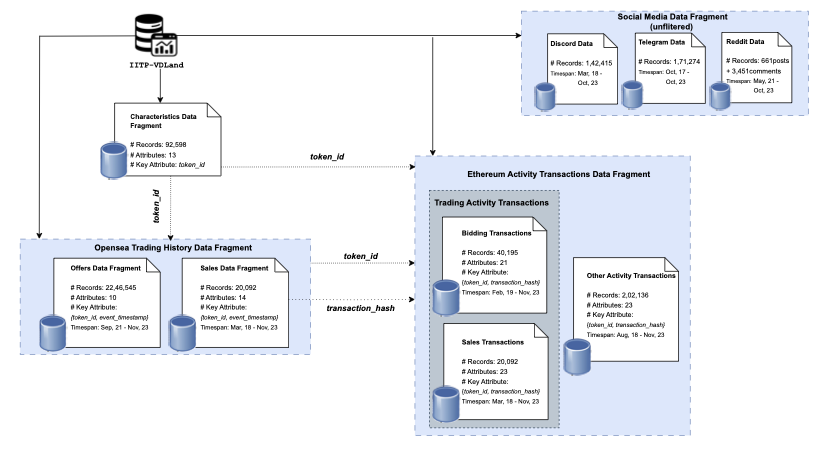}
\caption{Highlevel view of different data-fragments and relations}\label{highlevel_data}
\end{figure*}

A high-level view of various data-fragments and their relations in our dataset is depicted in Figure \ref{highlevel_data}. Let us now delve into the details of each of the fragments.

\subsection{Characteristics Data-Fragment}
Our data collection journey commenced with the utilization of the Decentraland API `\textit{tiles}' \cite{tiles} which furnishes us with a broad range of information covering essential characteristic attributes of total 92,598 virtual parcels. These attributes include unique Id, name, coordinates, owner-ID, etc. The proximity of a parcel to significant locations and landmarks within the Decentraland virtual world is a crucial aspect for many users, influencing its demand and supply. Keeping this in mind, we have gathered geographical proximity corresponding to these points of interest by using the Decentraland API `\textit{parcels}' \cite{parcels}. Furthermore, we have added an important parcel attribute, called Rarity score, which refers to the degree to which a particular digital asset is unique or scarce within a specific collection or series of NFTs \cite{Rarityscore}. While all parcels are inherently unique, their trade value can vary significantly based on this rarity factor. 

We measured the rarity score of our collections by applying one of the most popular rarity meter, namely Rarity.tools \cite{rarity}. It computes rarity score of individual trait (feature) in the collection and provides the resultant rarity as their sum, as per following:
\begin{equation*}
\scriptsize
\begin{split}
\textnormal{Rarity Score of a Trait Value} = \frac{\textnormal{Total number of items in collection}}
{\textnormal{Number of items with that Trait Value}}
\end{split}
\end{equation*}
Observe that Rarity.tools is suitable for our data as it also considers all none traits present, whereas many of the other rarity score meters \cite{Openrarity} tend to overlook none traits. Interestingly, the higher the Rarity score is, the rarer the asset is.

\subsection{OpenSea Trading History Data-Fragment}
In our relentless pursuit of historical trading data within Decentraland ecosystem, we use the OpenSea API `\textit{Get Events (by NFT)}' \cite{event}. This API allows us to delve deeper into the current market dynamics of parcels, gathering historical data about the bids and sales of parcels among the sellers and buyers. The process of trading Decentraland parcels commences with sellers listing their parcels on the marketplace, specifying the initial price. Then, interested buyers place bids during a specified auction duration and at the end of the auction, the highest bid wins the right to purchase the parcel. This entire sale process is facilitated by a local smart contract hosted on OpenSea network, which supports several tasks such as the transfer of ownership, release of payment, and verification. Therefore, we divided this trading data-fragment into two parts, as follow :

\subsubsection{Offers Dataset:} This critical data furnishes us with information about the offers proposed during the bidding process of parcels by the potential buyers. It is worthwhile to note that, while offering prices signify buyer interest, they don't guarantee successful sales. The dataset at hand is tied to OpenSea and exists within the off-chain domain of Ethereum. OpenSea employs a smart contract, named as `seaport', to empower users to offer prices for any parcels. OpenSea doesn't follow staking of bidding amount and hence save user from paying transaction fees and staking ether. The `confirmation of bid' is a wallet signature (EIP-712) necessary to make the process trustless. No one will be able to steal tokens by forging offers. This unique functionality of this seaport automatically execute sales when a user secures the highest bid. This off-chain dataset, encapsulates the dynamic interplay of user-initiated bids, the pricing mechanism, and the automated execution of sales, offering a far-reaching snapshot of trading activities within the OpenSea marketplace. As a result, there are a total of 22,46,545 offer records corresponding to 92,593 parcels spanning over a time period from September 04, 2021 to November 03, 2023.

\subsubsection{Sales Dataset:} This dataset offers a complete historical sales details of the parcels which are sold atleast once. As mentioned earlier, out of 92,598 parcels, only 9,220 parcels belong to this category comprising a total of 20,092 sales-records over a time period March 19, 2018 to November 03, 2023. This information plays a crucial role in several applications such as forecasting the prices of next sale of the parcel, finding out the probability of a parcel to be sold again, understanding temporal patterns of resale, average number of days it takes for the parcels to be resold, tracking price fluctuation over time, identifying recurring behaviors of any user or any particular parcel over a time period, etc. Since there exists a correspondence between the Offers dataset and the Sales dataset, one can relate these two datasets through the common attribute `token\_id' and `event\_timestamp'.  

\subsection{Ethereum Activity Transactions Data-Fragment}
As we delve into the transactions associated with Decentraland parcels, which operates on the Ethereum blockchain, our focus turn towards extracting relevant information from the transaction dataset available on Etherscan \cite{Etherscan} and Google BigQuery \cite{BigQuery}. Etherscan is a popular and widely used blockchain explorer and analytic platform specifically designed for the Ethereum blockchain. It provides a range of tools and services to explore, analyze, and interact with the Ethereum blockchain, making it an invaluable resource for Ethereum users, developers, researchers, and enthusiasts. Google BigQuery, on the other hand, is a high-performance and scalable platform for querying and analyzing vast datasets. To this aim, we consider the following two smart contract addresses  \cite{contract} deployed on the Decentraland mainnet, particularly linked with the parcels related services, and we extracted all relevant transactions from both Etherscan and Google BigQuery: (a) ERC721Bid smart contract at address `0xe479dfd9664c693b2e2992300\\930b00bfde08233', which facilitates auction-based transactions for parcels, allowing users to place bids, withdraw bids, and finalize auctions, and (b) LANDProxy smart contract at address~0xf87e31492faf9a91\-b02ee0deaad50d51d56d5d4d, which acts as an intermediary for parcel ownership and management within Decentraland, providing a standardized interface for interacting with parcels and enabling various functionalities such as transfers, estate creation, and access control. Our SQL query is meticulously crafted to target and extract intricate transaction details associated with the Decentraland ecosystem, utilizing addresses mentioned earlier. We subdivided our collected data into three distinct components: bids, sales and other associated transactions of the parcels. By segregating the data in this manner, we aim to facilitate targeted analysis of each facet of the trading ecosystem. This data fragment primarily contains transaction- and blockchain-specific information, including transaction-hash, block-number, log-index, gas consumption, services provided, etc. This acts as an important data resource for the development of AI tools for applications such as predicting users activity trends, correlation between gas price fluctuation and parcels' price dynamics, periods of increased or decreased transaction volume, anomaly detection, and many more.

\subsubsection{ERC721Bid Dataset:} This transaction dataset is the on-chain bids received by the parcels that serve as a real-time reflection of market sentiment and demand. The whole bidding process is initiated by invoking the `placeBid' function on the Bids-contract and the parcel owner has the authority to accept any active bids, provided they are not expired, and the bidder possesses sufficient amount for the transaction. To approve a bid, the asset owner needs to transfer the parcel to the Bids-contract using the `safeTransferFrom' function of the parcel contract, including additional data containing details of the accepted bid. Upon receiving the parcel, the Bids-contract verifies in its on `ERC721Received' function that the bid is still valid, and the bidder has adequate funds. If all conditions are met, the parcel transfers to the bidder, and the caller of the `safeTransferFrom' function is compensated with the declared amount from the bid. This dataset holds paramount importance in determining fair market values, establishing competitive pricing strategies, assessing bid success rates, gauging parcel popularity, and more. As a result, there are a total of 40,195 bid records spanning over a time period from February 26, 2019 to November 03, 2023.

\subsubsection{LANDProxy Dataset:} This dataset is extracted using the smart contract address `LandProxy'. It contains all the activities related to parcels ranging from create estate, transfer land, execute order, claim rewards and many more. In order to separate the sale dataset, the transaction\_hash of this dataset is matched with the sales dataset of \textit{OpenSea} to maintain the uniformity that includes 20,092 sales records of 9,220 parcels. This sale process is initiated through the `createOrder' function, once an order has been created on-chain, the asset will be listed for sale so that any user can buy it. This order is executed by calling the `executeOrder' function in the marketplace smart contract and then the exchange of amount and parcel take place between the two users. The rest 2,02,136 records are considered as other activities till November 03, 2023.

\subsection{Social Media Data-Fragment}
Social media platforms wield significant influence over Decentraland parcel pricing, shaping market dynamics through community engagement and generated hype. Platforms such as Discord \cite{Discord}, Telegram \cite{Telegram}, and Reddit \cite{Reddit}  serve as hubs for Decentraland enthusiasts, fostering discussions, sharing insights, and amplifying trends. This collective excitement surrounding specific collections significantly influences demand and, consequently, parcel prices. To analyze these pricing trends, data was gathered from these platforms' APIs spanning from October 20, 2017, to October 28, 2023. This yielded 317,801 pieces of content, including discussions, comments, and posts, offering valuable insight into parcel pricing influences.

Data from the Discord channels was exported as JSON files using the open-source application Discord Chat Exporter \cite{DataChatExporter}, covering the period from March 2018 to October 2023. The exported data consists of 142,415 records, including tags for the timestamp, author details, and message text.

Telegram data, covering October 2017 to October 2023, was extracted using the Pyrogram library which interacts with Telegram API and organized into three JSON files (Decentraland community messages, Mana-related messages, and Decentraland Market messages), totaling 171,274 records. To access the Telegram API, an application was registered on the Telegram website, providing the necessary credentials, including the api\_id and api\_hash. The target Telegram group's chat\_id, extracted from its URL, enabled focused data collection. This data includes message details, timestamps, and sender information. 

Reddit data, collected from the decentraland subreddit, provided valuable insights into community discussions. Utilizing the PRAW library to interact with the Reddit API, 661 posts and 3,451 comments were collected between May 2021 and October 2023. This process began with registering an application on the Reddit website to obtain the necessary API credentials. The dataset captures key elements including post IDs, titles, scores, URL, number of comments, selftext, and timestamps, compiled into both JSON and CSV files, named Redditcomments and SubRedditcomments, respectively. 

To comply with privacy requirements and prevent breaches of data protection regulations and intellectual property rights, we have implemented several precautionary measures. For all the social media data, sensitive information—including usernames, author IDs, and message IDs—has been \textit{anonymized} using a one-way, irreversible hash function. This ensures that the data cannot be reverted to its original form, thereby safeguarding the privacy of individuals involved.

\section{Exploratory Data Analysis}
\label{eda_app}

\begin{figure*}[!h]
    \centering
    \subfloat[Plaza Vs. Parcels]{\label{1}
		\includegraphics[width=0.3\linewidth]{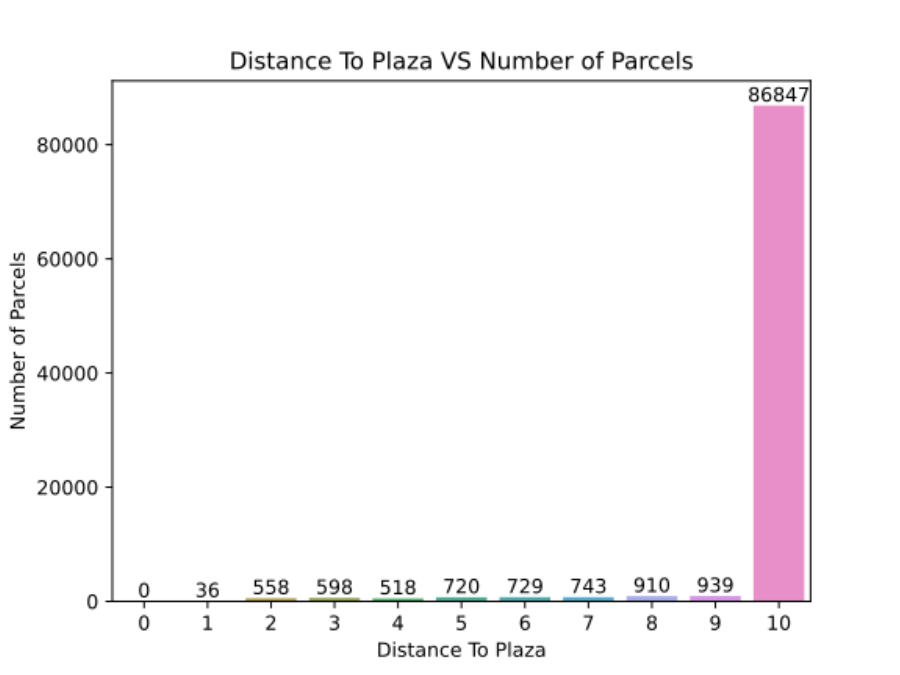}}
    \subfloat[District Vs. Parcels]{\label{2}
		\includegraphics[width=0.3\linewidth]{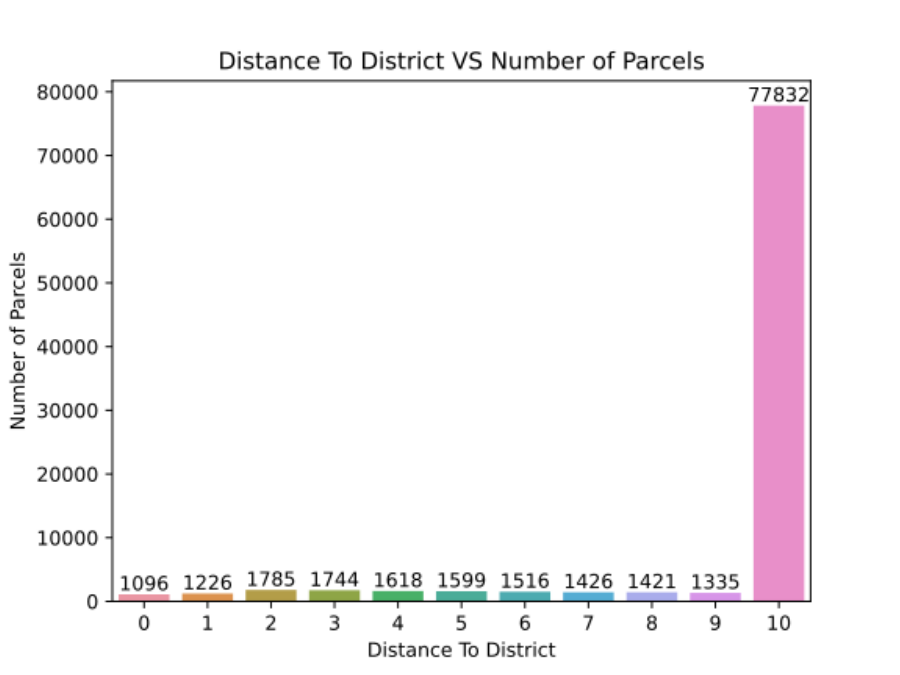}}
     \subfloat[Road Vs. Parcels]{\label{3}
		 \includegraphics[width=0.3\linewidth]{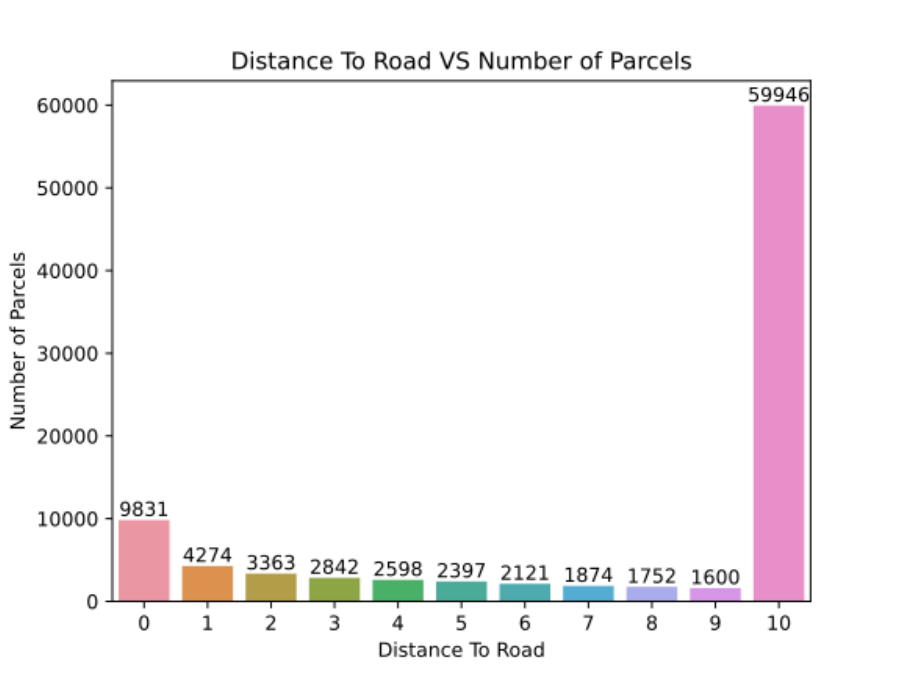}}

     \subfloat[Plaza Vs. Percentage]{\label{4}
	  \includegraphics[width=0.3\linewidth]{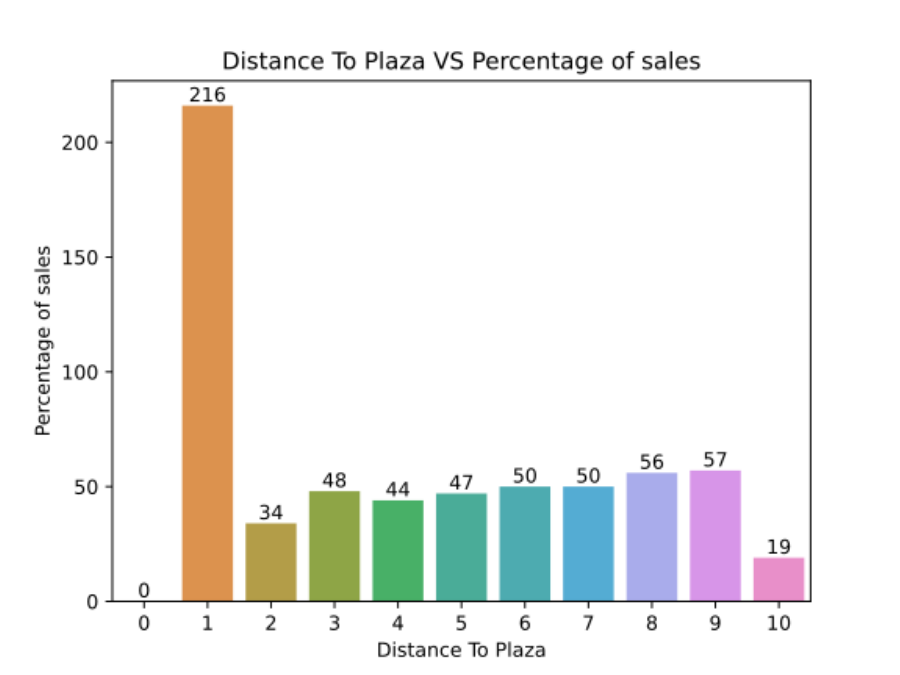}}
    \subfloat[District Vs.Percentage]{\label{fig:subfig4}
		  \includegraphics[width=0.3\linewidth]{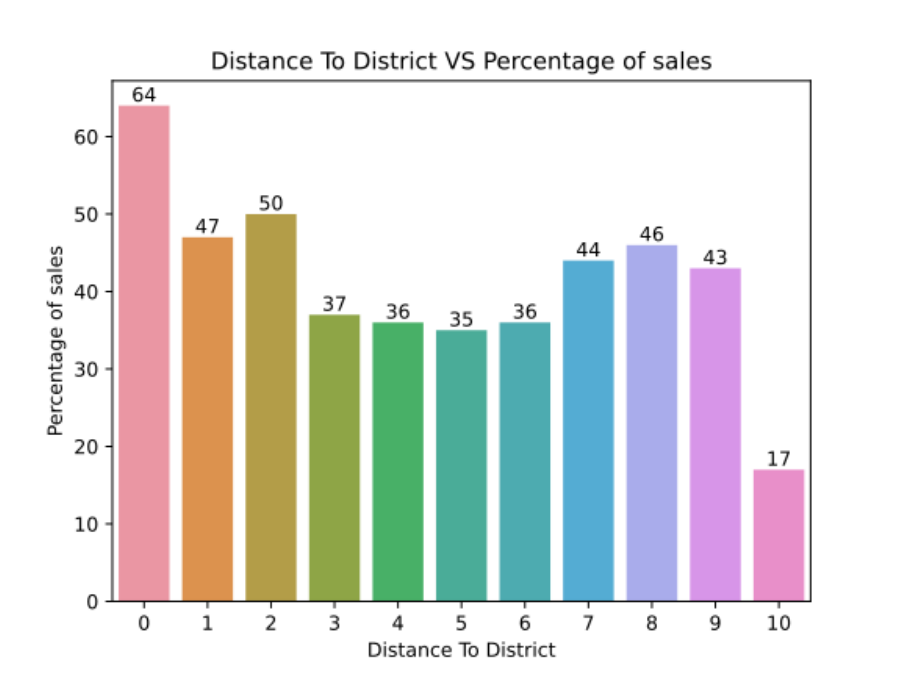}}
    \subfloat[Road Vs.Percentage]{\label{fig:subfig5}
		 \includegraphics[width=0.3\linewidth]{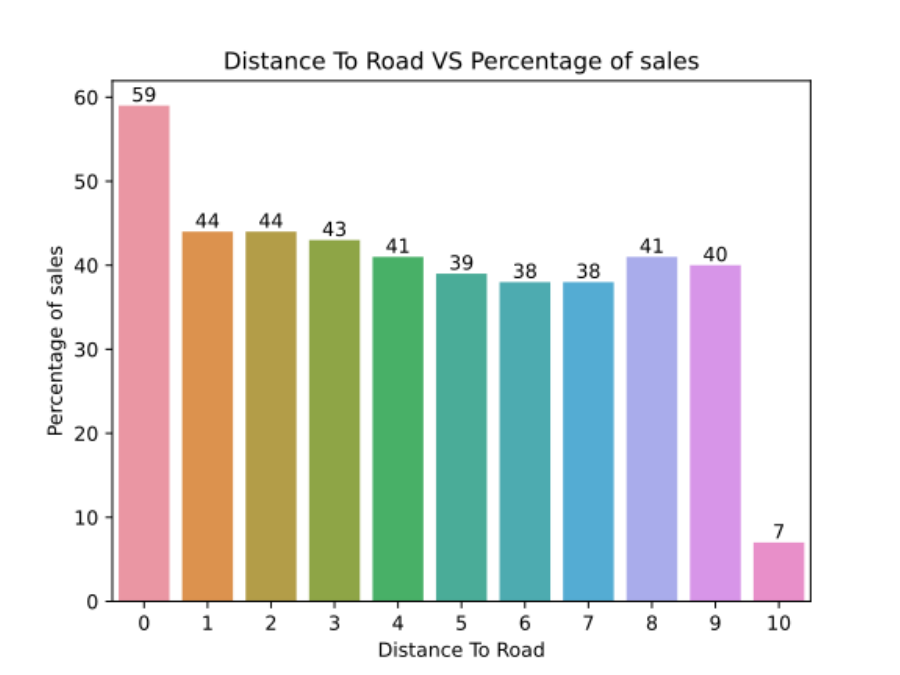}}	
\caption{Comparison of number of parcels and percentage of parcels sold with respective distances to geographical proximity}
\label{fig:DistanceComparision}
\end{figure*}

In this section, we provide an insightful information about the intricate patterns, relationships, and characteristics inherent in our dataset. As mentioned earlier in section \ref{sec:Decentraland}, the map of Decentraland is subdivided into different point of interests : Districts, Plazas, Roads, and Parcels. We analyze the relationship between the distances of parcels from the point of interests, exploring their impact on the number of sales. In Figure \ref{fig:DistanceComparision}, it is evident that the majority of parcels are situated at a considerable distance from their respective point of interests. However, when evaluating the percentage of sales, it indicates that users exhibit a tendency to acquire parcels in closer proximity to various landmarks resembling the real estate world. 

\begin{figure}[!h]
\centering
\includegraphics[width=\linewidth]{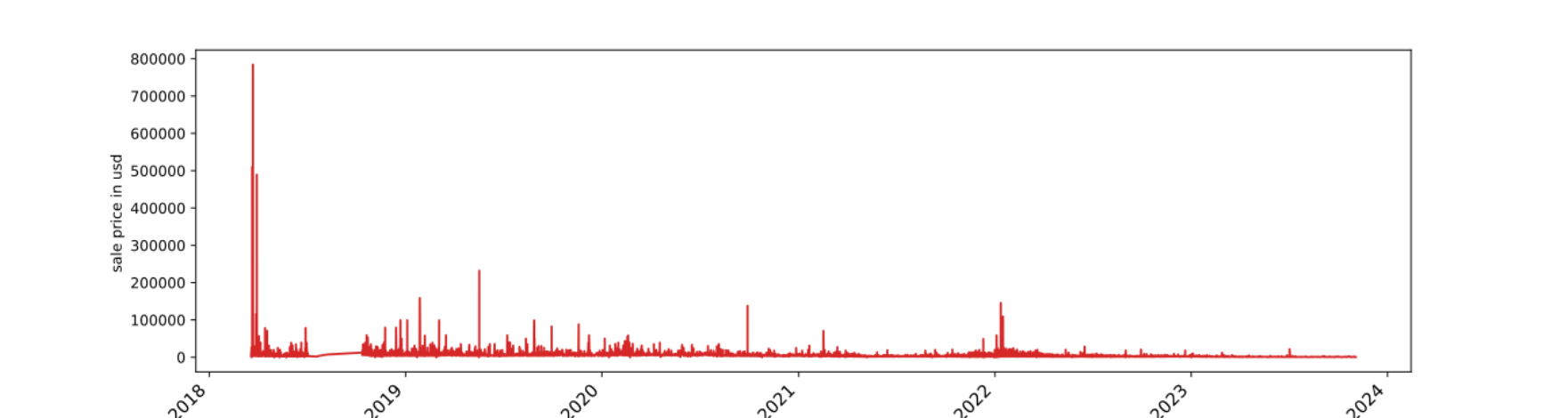}
\caption{Sale Price History of Decentraland parcels}
\label{fig:timeseries}
\end{figure}

In Figure \ref{fig:timeseries}, we conduct time series analysis on the value of parcels over the time. The highest sale value of \$784,124.0 was reached in 2018, but there is a constant decrease in the value of parcels throughout the subsequent years. It is likely to be influenced by a combination of market factors, platform developments, economic conditions, and changes in user behavior and preferences. 

\begin{figure}[!h]
      \centering
    \subfloat[Number of User Activities and Sales Vs Year]{\label{fig:activityVSsale}
		\includegraphics[width=0.45\linewidth]{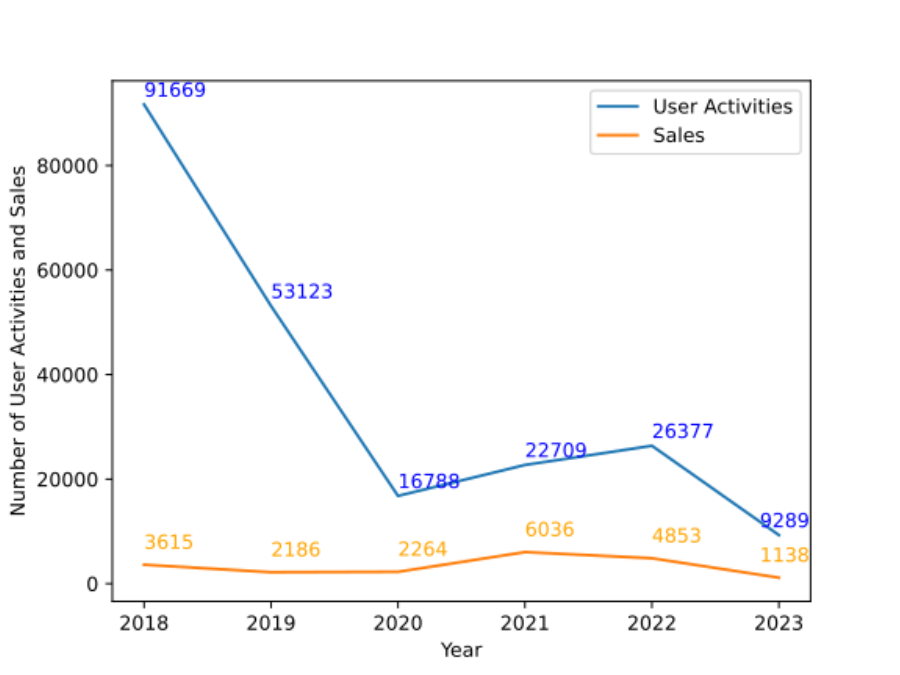}}
    \subfloat[Word-Cloud]{\label{wordcloud}
		\includegraphics[width=0.45\linewidth]{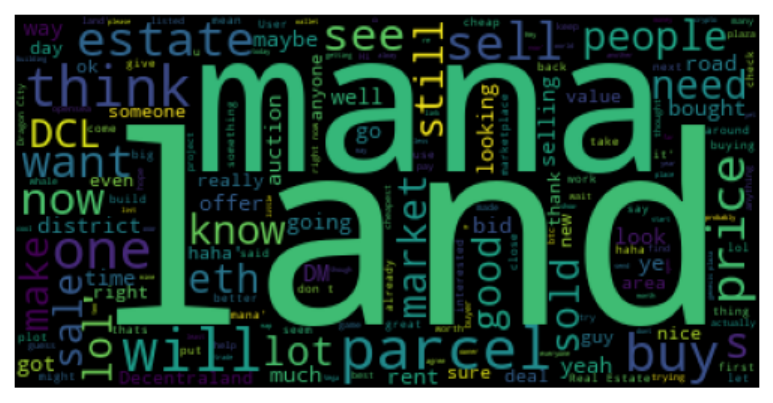}}
\caption{Impact of user activities and social media Decentraland wordcloud}
\label{fig:PeopleInterest}
\end{figure}

To understand the interest in Decentraland, we examined the user activities and sales over the years, extracting the significant terms centering around Decentraland. As shown in Figure \ref{fig:activityVSsale}, user activities sharply declined from 2018 to 2020, which impacted sales. However, during the pandemic (2020-2022), the surge in virtual worlds' popularity increased demand for parcels, boosting Decentraland's valuation. In 2023, as the world reopened, interest in buying parcels waned, leading to a decline in demand, reflecting high volatility. Figure \ref{wordcloud} highlights how social media discussions, particularly on Discord, significantly influence parcel pricing by driving enthusiasm and hype.
Our dataset reveals that 100 seller addresses initiate transactions with nearly 8,280 recipient addresses, offering a unique opportunity to study network dynamics and interconnections, similar to \cite{nadini2021mapping}, but focused on Decentraland parcels. Of the parcels, 4,150 were sold only once between 2018-2023, while 5,070 were resold multiple times, indicating a need for further analysis on resale patterns. While numerous potential analyses could yield valuable insights, we concentrate on practical applications of our dataset, employing machine- and deep-learning methods for tasks like price prediction and binary parcel classification to showcase its significance within the Decentraland ecosystem. 

\section{Benchmark Evaluation: Parcels valuation and pricing}\label{sec:benchmark}
In our benchmark, we consider a diverse range of state-of-the-art price prediction models, encompassing both regression and classification tasks across various NFTs and cryptocurrencies. A brief summary of these models are mentioned in \ref{tab_ds7}. Our primary objective is to utilize these baseline models to evaluate their performance on our proposed dataset. While we adopt regression task aiming at predicting the price of Decentraland parcels, the classification task enables us to assess the likelihood of parcels for being resold. The experiments are conducted on a system equipped with Intel core i5 $10^{th}$ gen CPU with Windows 10 Pro operating system, 16GB of RAM and a 256GB SSD. 

\begin{table*}[ht!]
\centering\scriptsize
\begin{tabular}{|p{0.7in}|p{0.2in}|p{2.2in}|p{1.5in}|}
\hline
{\bf State-of-the-Art proposals}  & {\bf Year} & {\bf Brief Summary} & {\bf ML/DL models used} \\ \hline
Henriques et al. \cite{henriques2023forecasting}    & 2023     & Forecasted NFT coin prices using machine learning approach.              &    Random Forest, Extremely Randomized Trees, Tree Boosting, SVM, Lasso, Naive Bayes     \\ \hline
Seabe et al. \cite{seabe2023forecasting}    & 2023     & Forecasted cryptocurrency prices using deep learning approach.              & LSTM, GRU, Bi-Directional LSTM     \\ \hline
Z. wang et al. \cite{wang2023prediction}   & 2023     & Predicted NFT sale price fluctuations on OpenSea.              &  AdaBoost, Random Forest     \\ \hline
Brunet et al. \cite{brunet2023exploring}    & 2023     & Explored different Metaverse platforms and created an analysis tool using machine learning model.              & Spearman correlation, XGBoost    \\ \hline
Dawod et al. \cite{dawod2023nft}      & 2023      & Evaluated machine learning algorithms to appraise NFT real-price based on characteristics, market event information, and rarity score.              &     LightGBM, CatBoost, XGBoost, RF, TabNet, Polynomial, LR, SVM, Ridge, Lasso, ElasticNet    \\ \hline
J. Luo et al. \cite{luo2023understanding}      & 2023     & Demonstrated NFT price movement through tweets keywords analysis.              & Support vector machine, multilayer perceptron, Transformer
        \\ \hline
Ghosh et al. \cite{ghosh2023prediction}   & 2023      & Predicted and interpreted NFT and DeFi prices through ensemble machine learning.             & Isometric mapping - GBR, Uniform manifold approximation and projection - RF, Decision Tree, Support Vector Regressor, ARIMA, SARIMA
        \\ \hline        
Kin-Hon-Ho et al. \cite{ho2022analysis}    & 2022     & Highlighted the intricate interplay between rarity and utility in determining prices of play-to-earn gaming NFTs on Axie Infinity.                & XG BOOST        \\ \hline
Branny et al. \cite{branny2022non}    & 2022     &Forecasted NFT sale prices using multiple multivariate
time series datasets containing features related to the NFT
market space.              & Linear Regression, Decision Tree, Bayesian Ridge, LSTM
        \\ \hline
Kapoor et al. \cite{kapoor2022tweetboost}       & 2022     & Predicted NFT asset value with Twitter and OpenSea Interaction Analysis.              & Logistic Regression, SVM, Random Forest, Light gbm, XG Boost       \\ \hline
Jain et al. \cite{jain2022nft}      & 2022      & Examined the correlation between NFT valuation and various features: market data, metadata and social trends data.               & RNN, Linear Regression        \\ \hline
Nadini et al. \cite{nadini2021mapping}      & 2021     & Analyzed market trades to predict NFT sales using simple machine learning approach.             & Linear Regression, Adaboost, PCA, AlexNet        \\ \hline
Chen et al. \cite{chen2020bitcoin}      & 2020      &Considered the sample’s granularity and feature dimensions for Bitcoin price prediction . &Logistic Regression, Random Forest, XGBoost, LSTM, SVM    \\ \hline
JAY et al. \cite{jay2020stochastic}     & 2020     & Introduced a stochastic module to capture markets' reaction and observed feature activations of neural networks to stimulate market volatility.              &     MLP, LSTM    \\ \hline
Felizardo et al. \cite{felizardo2019comparative}      & 2019     & Comparative study of Bitcoin price prediction using different machine learning approach.               &     ARIMA, SVM, Random Forest, LSTM, WaveNets    \\ \hline
Saad et al. \cite{saad2019toward}       & 2019     &  Built a machine-learning model to analyze the market of cryptocurrency for highly accurate prediction.             & Linear Regression, Random Forest, Gradient Descent    \\ \hline
Lahmiri et al. \cite{lahmiri2019cryptocurrency}    & 2019      & Detected chaos and fractal characteristics along with LSTM to predict price of cryptocurrencies using the largest Lyapunov Exponent (LLA) and Detrended Fluctuation Analysis (DFA).   & LSTM, GRNN, LLE, DLNN    \\ \hline
Mcnally et al.\cite{mcnally2018predicting}     & 2018      & Predicted price of Bitcoin in USD.             &     RNN, LSTM, Arima    \\ \hline
Laura et al. \cite{alessandretti2018anticipating}      &2018      & Anticipated the short-term evolution of the cryptocurrency market using two ensemble models of regression trees.               & XGBoost, RNN, LSTM     \\ \hline
JANG et al. \cite{jang2017empirical}     & 2017      & Predicted Bitcoin price using BNN, based on blockchain information.             & BNN, Linear Regression   \\ \hline
Sin et al. \cite{sin2017bitcoin}       & 2017      & Explored the dependencies of next day price on Bitcoin features.              &     MLP, GASEN    \\ \hline
\end{tabular}%}
\caption{Brief summary of the state-of-the-art price prediction models for NFTs and cryptocurrencies}
\label{tab_ds7}
\end{table*}

\subsection{Data Preprocessing}
In order to evaluate the models on our proposed dataset, we seamlessly integrate the characteristics and OpenSea sales data fragment based on `token\_id' and `last sale price' of parcels, with the latter being designated as the prediction target. From the total 92,598 parcels, only 9,220 parcels are identified as being sold at least once. These selected parcels are then merge with the Ethereum activity transactions data fragment, utilizing the `transaction\_hash' to incorporate the gas price incurred during the trading. The parcel price and gas price associated with each transaction are converted to USD, shedding light on the effective transaction costs incurred in the buying and selling of parcels. To streamline our experiments, we exclusively retain numerical features, eliminating irrelevant attributes containing text, links, IDs, and dates. The refined set of features primarily includes coordinates, geographical proximity to the points of interest, exchange rates in USD and Ether for the payment tokens, gas fees, calculated rarity scores and the most recent sale price of the parcels. The prices of parcels in our dataset exhibit significant volatility. To mitigate the influence of outliers, we adopt the outlier removal methodology outlined in \cite{kapoor2022tweetboost}. Specifically, we categorize the dataset into four tiers based on asset value: `Extreme' (greater than \$100,000), `High' (exceeding \$10,000), `Mid' (falling between \$1,000 and \$10,000), and `Low' (below \$1,000). This results 147 assets in the low-value category, 8,391 in the mid-value range, 673 classified as high-value assets, and 9 identified as extreme-value assets. Subsequently, we opt to disregard extreme-, high-, and low-value assets as outliers, focusing solely on mid-value assets as a more representative set of the data. Following this, we perform normalization of the target price through Min-Max scaling \cite{saad2019toward,felizardo2019comparative}, according to the following equation:
\begin{align*}
     X_{scaled} = X_{std} * (max-min) + min
\end{align*}
where
\begin{equation}
    X_{std} = {\frac{X - X.min(axis=0)}{X.max(axis=0) - X.min(axis=0)}}
\end{equation}

\noindent Furthermore, motivated from the findings in \cite{ghosh2023prediction,smuts2019drives}, we expand our analysis by integrating crucial external data, which act as economic indicators \cite{investing,GoogleTrend}, including Ether-, Bitcoin-, Crude oil-, Gold-prices and Google trend, which has already been proven to have a significant impact on the NFT marketplace. This strategic feature selection process aims to maintain focus and potentially enhance the accuracy of predictions by aligning with the critical factors influencing pricing dynamics within our dataset.

\subsection{Performance Metrics}
As the target variable is price, a continuous variable, we have chosen several regression evaluation metrics to validate our experimental results \cite{error_metric}. These include Mean Absolute Error (MAE), Mean Squared Error (MSE), Root Mean Squared Error (RMSE), $R^2$ score, Root Mean Squared Log Error (RMSLE), and Mean Absolute Percentage Error (MAPE). The classification metrics to validate our experimental results include Accuracy, Precision, Recall, and F1 score \cite{error_metric}. Let \textit{TP}, \textit{TN}, \textit{FP}, and \textit{FN} denote true positve, true negative, false positive, and false negative respectively. 

\subsection{Training and Performance Evaluation}
In this section, we present the experimental evaluation results obtained by applying the state-of-the-art regression and classification approaches on our dataset. The aim is to assess their predictive performance and determine which models are best suited for our dataset. To conduct this evaluation, we randomly divided the dataset into two parts: a training set containing 80\% of the parcels and a test set containing remaining 20\% parcels. 

Among the machine learning models applied for regression analysis, we observe that the Extra Trees Regressor, Light Gradient Boosting Machine, Random Forest Regressor, and Extreme Gradient Boosting consistently exhibit superior performance across multiple metrics, as depicted in Table \ref{tab_regression}. Specifically, the Extra Trees Regressor achieves an $R^2$ score of 0.8251, exhibiting the highest level of explanatory power among the models tested. The other ensemble models Light Gradient Boosting Machine, Random Forest Regressor, and Extreme Gradient Boosting follow closely with $R^2$ score of 0.8211, 0.8195, and 0.8151 respectively showcasing their effectiveness in explaining the variance in the target variable. The underlying reason of Extra Trees Regressor performing better as compared to other models lies in its unique approach of randomly selecting subsets of features, leading to increase diversity among trees that helps to reduce variance and overfitting. We also find that Gradient Boosting and Decision Tree gives $R^2$ score of 0.7364 and 0.6474 respectively that is relatively less than other ensemble models. These models while less susceptible to overfitting, still relies on the sequential addition of weak learners, which does not fully capture the underlying relationships in the data. Conversely, linear models such as Linear regression, Ridge regression, Lasso regression, Bayesian Ridge regression, Elastic Net, and AdaBoost exhibit comparatively poor performance across the evaluated metrics with $R^2$ score lying around 0.36, due to the volatility and non-linearity inherent in our dataset. 

\begin{table*}[t]
\centering\scriptsize
\begin{tabular}{|p{1in}|l|l|l|l|l|l|l|}
\hline
{\bf Model} & {\bf MAE}     & {\bf MSE}       & {\bf RMSE}       & {\bf R2}      & {\bf RMSLE}   & {\bf MAPE}       \\
\hline
{\bf Extra Trees Regressor} \cite{henriques2023forecasting}     & {\bf 549.9695}	& {\bf 755238.1578}	& {\bf 868.6787}	& {\bf 0.8251}	& {\bf 0.2309}	&{\bf 0.1678}	 \\ \hline
Light Gradient Boosting Machine \cite{dawod2023nft, kapoor2022tweetboost} &591.9959	&772802.0429	&878.8025	&0.8211	&0.2375	&0.1813	   \\ \hline
Random Forest Regressor \cite{henriques2023forecasting, wang2023prediction, ghosh2023prediction, kapoor2022tweetboost, chen2020bitcoin, felizardo2019comparative, saad2019toward, dawod2023nft}         &565.7065	&779215.1271	&882.0563	&0.8195	&0.2357	&0.1726	  \\ \hline
Extreme Gradient Boosting \cite{brunet2023exploring, dawod2023nft, ho2022analysis, kapoor2022tweetboost, chen2020bitcoin, alessandretti2018anticipating}     & 588.1852	&798352.8312	&892.8080	&0.8151	&0.2424	&0.1807	   \\ \hline
        
Gradient Boosting Regressor \cite{ghosh2023prediction}     & 752.5699	&1139456.3378	&1066.7778	&0.7364	&0.2886	&0.2382	   \\ \hline
Decision Tree Regressor \cite{ghosh2023prediction, branny2022non}      & 724.2672	&1521704.7801	&1232.1022	&0.6474	&0.3154	&0.2153	   \\ \hline
Linear Regression  \cite{nadini2021mapping, branny2022non, saad2019toward, jang2017empirical, dawod2023nft}           & 1257.0171	&2742406.8567	&1655.4335	&0.3653	&0.4605	&0.4407	   \\ \hline
Ridge Regression  \cite{dawod2023nft}              & 1260.2644	&2746626.7865	&1656.7530	&0.3643	&0.4607	&0.4410	   \\ \hline
AdaBoost Regressor \cite{nadini2021mapping, wang2023prediction}           & 1469.5244	&2753052.8910	&1658.5956	&0.3628	&0.5359	&0.6201	
         \\ \hline
Lasso Regression  \cite{dawod2023nft, henriques2023forecasting}              & 1260.8927	&2753645.3368	&1658.8414	&0.3627	&0.4609	&0.4412	   \\ \hline
Bayesian Ridge  \cite{branny2022non}                & 1263.4940	&2765753.7330	&1662.4404	&0.3600	&0.4619	&0.4426	   \\ \hline
Elastic Net \cite{dawod2023nft}                   & 1263.4951	&2765622.9268	&1662.4024	&0.3600	&0.4619	&0.4427	  \\ \hline
\end{tabular}
\caption{Performance Analysis of Regression Models} 
\label{tab_regression}
\end{table*}  

We also observe a similar trend in the performance of the machine learning based classification approaches, as depicted in Table \ref{tab_classification}. Notably, the Extra Trees Classifier stands out with the highest accuracy of 74.23\%, and Random Forest Classifier model excel with an accuracy of 73.74\%, achieving the highest Recall (81.25\%), and F1 score (76.90\%). These models indicate their ability to correctly classify instances across different classes while minimizing false positives and false negatives. Despite its slightly lower accuracy of 71.40\%, the Light Gradient Boosting Machine exhibits a superior precision at 73.48\%. Other models such as Extreme Gradient Boosting, Gradient Boosting Classifier, and Decision Tree also demonstrate commendable performance, with accuracies of 71.35\%, 66.60\%, and 64.46\% respectively. Linear models like Adaboost, Ridge, Logistics Regression, and SVM with a linear kernel achieve satisfactory accuracy levels, with varying performances across different metrics. However, it is crucial to consider the trade-offs between different metrics base on the specific goals of the classification task. Overall, these results highlight the effectiveness of ensemble methods such as the Extra Trees in accurately predicting the target variable in our dataset over the linear models.

\begin{table*}[t]
\centering\scriptsize
\begin{tabular}{|p{2in}|p{0.6in}|p{0.6in}|p{0.6in}|p{0.6in}|}
\hline
{\bf Model} & {\bf Accuracy}     & {\bf Recall}       & {\bf Prec}       & {\bf F1}   \\
\hline
{\bf Extra Trees Classifier} \cite{henriques2023forecasting}                 &{\bf 0.7423}	&0.8021	&0.7284	&0.7634	    \\ \hline
Random Forest Classifier \cite{henriques2023forecasting, wang2023prediction, ghosh2023prediction, kapoor2022tweetboost, chen2020bitcoin, felizardo2019comparative, saad2019toward, dawod2023nft}      & 0.7374	&\textbf{0.8125}	&0.7303	&\textbf{0.7690}  \\ \hline
Light Gradient Boosting Machine \cite{dawod2023nft, kapoor2022tweetboost} & 0.7140	&0.8072	&\textbf{0.7348}	&0.7646	  \\ \hline
Extreme Gradient Boosting \cite{brunet2023exploring, dawod2023nft, ho2022analysis, kapoor2022tweetboost, chen2020bitcoin, alessandretti2018anticipating}     & 0.7135		&0.7744	&0.7297	&0.7512  \\ \hline
Gradient Boosting Classifier\cite{ghosh2023prediction}    & 0.6660		&0.7921	&0.6705	&0.7261    \\ \hline
Decision Tree Classifier \cite{ghosh2023prediction, branny2022non}        &0.6446	&0.6658	&0.6884	&0.6767	   \\ \hline
Ada Boost Classifier \cite{nadini2021mapping, wang2023prediction}        & 0.6349	&0.7455	&0.6517	&0.6952	   \\ \hline
Ridge Classifier\cite{dawod2023nft}                & 0.6101	&0.7717	&0.6218	&0.6886	  \\ \hline
Logistic Regression  \cite{kapoor2022tweetboost, chen2020bitcoin}          & 0.6011	&0.7674	&0.6146	&0.6824  \\ \hline
SVM - Linear Kernel \cite{henriques2023forecasting, dawod2023nft, luo2023understanding, kapoor2022tweetboost, chen2020bitcoin, felizardo2019comparative}           & 0.5650	&0.7353	&0.5342	&0.6119	   \\ \hline
\end{tabular}
\caption{Performance Analysis of Classification Models} 
\label{tab_classification}
\end{table*}
Let us now consider the state-of-the-art deep learning models, including Long Short-Term Memory (LSTM), Recurrent Neural Network (RNN), and Multilayer Perceptron (MLP). In order to apply them on our dataset, we apply a standardized configuration across the board. This configuration consist of an input layer with 256 neurons, followed by two hidden layers with 128 and 64 neurons respectively. We utilize the Adam optimizer for optimization, with the Rectified Linear Unit (ReLU) serving as the activation function. MSE is employed as the loss function for regression task, while Binary Crossentropy is used for classification purpose. Setting the learning rate to 0.001, batch size to 32, and training epochs to 50 ensure that our models refine their parameters without overfitting or excessive computational costs. The evaluation results are shown in Table \ref{trtt}. Interestingly, our analysis reveals a spectrum of performance among the deep learning models, ranging between traditional linear models and ensemble-based approaches. Notably, RNN achieves the highest $R^2$ score of 0.7466, closely followed by LSTM of 0.7410, and MLP of 0.7210 for regression task. For classification task, MLP outperforms the others with an accuracy of 66\% . These findings indicate a balanced performance across the models, demonstrating their capacity to capture intricate patterns within the data.

\begin{table*}[t]
\centering
\scriptsize
\subfloat[Classification Models]{\label{subtable1}
\begin{tabular}{|l|c|c|c|c|}
\hline
{\bf Model} & {\bf Acc} & {\bf Recall} & {\bf Prec} & {\bf F1} \\ \hline
\textbf{MLP} \cite{sin2017bitcoin, luo2023understanding, jay2020stochastic} & \textbf{0.660} & \textbf{0.721} & \textbf{0.698} & 0.664 \\ \hline
RNN  \cite{alessandretti2018anticipating, mcnally2018predicting, jain2022nft} & 0.654 & 0.687 & 0.684 & \textbf{0.683} \\ \hline
LSTM \cite{felizardo2019comparative, alessandretti2018anticipating, lahmiri2019cryptocurrency, mcnally2018predicting, jay2020stochastic, chen2020bitcoin, branny2022non, seabe2023forecasting} & 0.619 & 0.674 & 0.652 & 0.667 \\\hline
\end{tabular}
}
\vspace{0.2in}
\subfloat[Regression Models]{\label{subtable2}
\begin{tabular}{|l|c|c|c|c|c|c|}
\hline
{\textbf Model} & {\textbf MAE} & {\textbf MSE} & {\textbf RMS\-E} & {\textbf R2} & {\textbf RMS\-LE} & {\textbf MAP\-E} \\\hline
\textbf{RNN} \cite{alessandretti2018anticipating, mcnally2018predicting, jain2022nft} & \textbf{701\newline.231} & \textbf{1117\newline.524} & \textbf{1057\newline.130} & \textbf{0.746} & \textbf{0.273} & \textbf{0.199} \\ \hline
LSTM \cite{felizardo2019comparative, alessandretti2018anticipating, lahmiri2019cryptocurrency, mcnally2018predicting, jay2020stochastic, chen2020bitcoin, branny2022non, seabe2023forecasting} & 735\newline.587 & 1141\newline.768 & 1068\newline.535 & 0.741 & 0.284 & 0.224 \\ \hline
MLP \cite{sin2017bitcoin, luo2023understanding, jay2020stochastic} & 730\newline.039 & 1217\newline.754 & 1100\newline.797 & 0.721 & 0.288 & 0.216  \\\hline
\end{tabular}
}
\caption{Performance Results of Deep Learning Models}
\label{trtt}
\end{table*}

\subsubsection{Analysing Feature Importance:}
\begin{figure*}[h]
       \centering
 	   \subfloat[Regression]{
          \label{fig:subfig7}
		\includegraphics[width=0.5\linewidth]{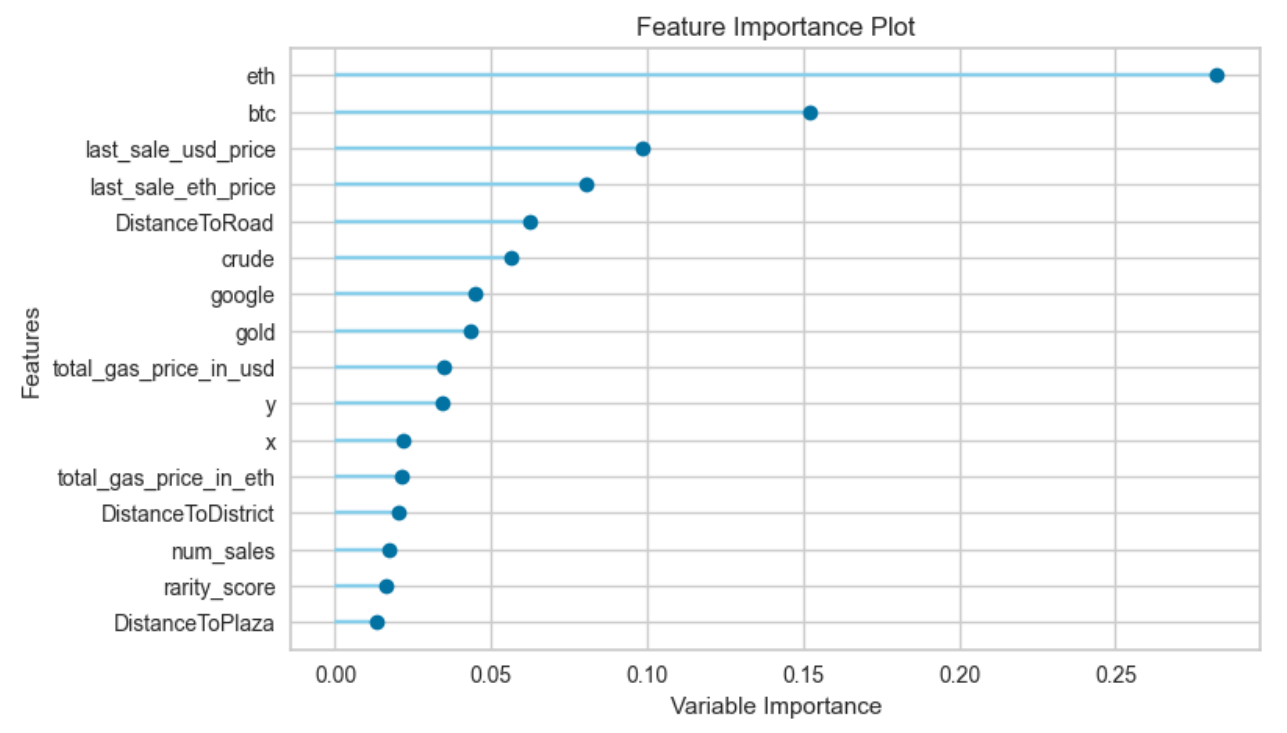}
    }
    \subfloat[Classification]{        \label{fig:subfig8}
		\includegraphics[width=0.5\linewidth]{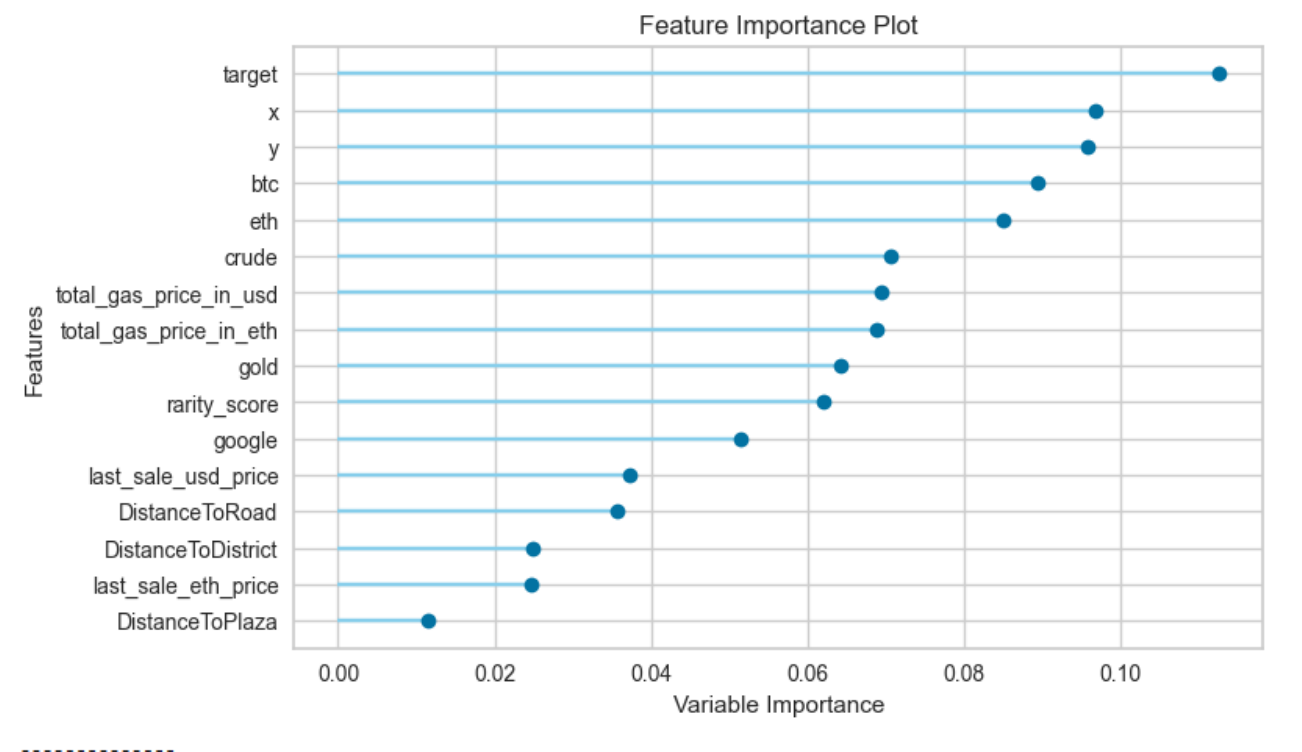}
    }
    \caption{Feature Importance}
    \label{fig:subfigureall9}
\end{figure*}

In this section, we analyze the importance of various features under consideration, which contribute to the overall performance in both the tasks for Extra Trees models. As we can see in Figure \ref{fig:subfig7}, for regression task, the prices of ether and bitcoin alongside their exchange-rates heavily influence the price of the parcel, thereby underscoring the interconnectedness of the cryptocurrency market, as highlighted in \cite{ghosh2023prediction}. In addition, while the attribute `DistanceToRoad' influences the sale prices of the parcels (Figure \ref{fig:subfig7}), the parcels coordinates (x, y) play a pivotal role in assessing the likelihood of a parcel being resold (Figure \ref{fig:subfig8}), as reported in \cite{brunet2023exploring}. Rarity score proves to be a strong predictor in classification task but shows reduced efficacy in regression. Gas prices do not significantly contribute to the price prediction but exhibit notable significance in parcel resale forecasts. Interestingly, google trend, crude oil- and gold-prices also make moderate contributions to both predictions. It is worth mentioning that other geographical proximities, such as `DistanceToPlaza' and `DistanceToDistrict', perform poorly in both of the tasks. This could be attributed to most parcels being far from plazas and districts, with a value of 10 as seen in Figure \ref{fig:DistanceComparision}, thereby diluting their significance. These findings demonstrate the potential to analyze and predict market trends of Decentraland parcels. With the appropriate data and organization, researchers can gain valuable insights into the usage patterns and operational dynamics of this platform.

\section{Discussions and Future Research Scope}\label{sec:Discussion}

The Metaverse, hailed as the future of the internet, is experiencing a surge in virtual land trading, particularly on the Decentraland platform \cite{Decentraland}. While prior research has explored Decentraland NFT pricing \cite{dowling2022fertile, goldberg2021land, yencha2023spatial, nadini2021mapping, dowling2022non, schonbaum2022decentraland, pinto2022nft, kaneko2021time, luo2023understanding, costa2023show, brunet2023exploring, luo2023unveiling}, existing datasets lack the breadth and depth needed for comprehensive analysis.

This paper introduces \textsf{IITP-VDLand} , a comprehensive five-year dataset of Decentraland parcel data. Sourced from platforms like Decentraland \textit{Decentraland} \cite{Decentraland}, \textit{OpenSea} \cite{OpenSea}, \textit{Etherscan} \cite{Etherscan}, \textit{Google BigQuery} \cite{BigQuery}, \textit{Discord} \cite{Discord}, \textit{Telegram} \cite{Telegram}, and \textit{Reddit} \cite{Reddit}, \textsf{IITP-VDLand} includes novel features such as a parcel Rarity score, detailed blockchain transaction data , and social media sentiment. This dataset is organized into four fragments: Characteristics, OpenSea Trading History, Ethereum Activity Transactions, and Social Media Data.

We've discovered a correlation between parcel sale prices and their proximity to points of interest, mirroring real-world trends. Additionally, our analysis reveals a direct link between the volume of activities and the number of sales. This rich dataset holds potential for applications such as analyzing transaction costs, market liquidity, and pricing trends, to name a few. For now, we focus on showcasing its capabilities through regression and classification approaches using over 20 machine and deep learning models. The Extra Trees algorithm showed strong performance $R^2$ score of 82.51\%  for price prediction and 74.23\% accuracy for resale prediction, highlighting the dataset's complex nature.

Our research paves the way for a broader exploration of the Decentraland ecosystem, including aspects like avatars, wearables, and other virtual experiences. Predicting trends in visual and textual features could leverage techniques like ARIMA and neural network models such as CNNs and GNNs. The multimodal nature of our dataset (textual, visual, and social) allows for enhanced precision through multimodal learning techniques. Beyond price prediction, the rich attributes within our dataset open doors to numerous research avenues. These include sentiment analysis, understanding network dynamics within trader behavior, and tracking complex trends within the Decentraland ecosystem.

\section{Acknowledgement}
This research is partially supported by the research grant provided by IIT Bhilai Innovation and Technology Foundation (IBITF).

\section{Data availability}
The complete dataset is available for download at: \url{https://dx.doi.org/10.21227/qv8s-7n53}

%\printbibliography %Prints bibliography

\bibliographystyle{plainurl}
\bibliography{References.bib}

\end{document}